\documentclass[twoside,11pt]{article}
\usepackage{pgm}
\usepackage{tikz}
\usetikzlibrary{positioning,shapes.geometric}
\usepackage{dsfont,bm}
\usepackage{bbm}
\usepackage{amsmath}
\usepackage[justification=centering]{caption}
\usepackage{amssymb} 
\usepackage{amsthm} 
\usepackage{etoolbox}
\usepackage{floatrow}
\usepackage[justification=centering]{caption}
\newfloatcommand{capbtabbox}{table}[][\FBwidth]
\usepackage{algpseudocode}
\usepackage{algorithm}
\usepackage{hyperref}
\DeclareMathOperator*{\argmin}{argmin}
\usepackage{amsthm}
\usepackage{gensymb}
\newtheorem{proposition}{Proposition}
\setcitestyle{authoryear,open={(},close={)}}
\usepackage[threshold=0]{csquotes}

\usepackage{hyperref,color,soul,booktabs}
\setulcolor{blue}
\newcommand{\BibTeX}{\textsc{B\kern-0.1emi\kern-0.017emb}\kern-0.15em\TeX}

\ShortHeadings{Causal Feature Learning for Utility-Maximizing Agents}{}

\begin{document}

\title{Causal Feature Learning for Utility-Maximizing Agents}

\author{\Name{David Kinney} \Email{david.kinney@santafe.edu}\\
    \addr{Santa Fe Institute}
    \and
   \Name{David Watson} \Email{david.watson@oii.ox.ac.uk}\\
  \addr{Oxford Internet Institute}}

\maketitle

\begin{abstract}
Discovering high-level causal relations from low-level data is an important and challenging problem that comes up frequently in the natural and social sciences. In a series of papers, Chalupka et al.\ (\citeyear{Chalupka2015VisualCF}, \citeyear{Chalupka:2016:UDE:3020948.3020957}, \citeyear{chalupka2016multi}, \citeyear{chalupka2017causal}) develop a procedure for \textit{causal feature learning} (CFL) in an effort to automate this task. We argue that CFL does not recommend coarsening in cases where pragmatic considerations rule in favor of it, and recommends coarsening in cases where pragmatic considerations rule against it. We propose a new technique, \textit{pragmatic causal feature learning} (PCFL), which extends the original CFL algorithm in useful and intuitive ways. We show that PCFL has the same attractive measure-theoretic properties as the original CFL algorithm. We compare the performance of both methods through theoretical analysis and experiments.
\end{abstract}
\begin{keywords}
Causal Feature Learning; Coarse-Graining; Bayesian Networks; Expected Utility.
\end{keywords}

\section{Introduction}
In many scientific contexts, one goal of inquiry is to discover types of fine-grained events that can be grouped together into a smaller set of more coarse-grained events. This is especially true in causal analysis, where the goal is often to find some hierarchical structure to improve both the tractability and the representational adequacy of models. Developing efficient and reliable computational methods for mapping low-level data to high-level phenomena is therefore an important step in the more general task of automating, at least partially, the process of scientific discovery.\par 

Chalupka et al.\ (\citeyear{Chalupka2015VisualCF}, \citeyear{Chalupka:2016:UDE:3020948.3020957}, \citeyear{chalupka2016multi}, \citeyear{chalupka2017causal}) propose a procedure called \textit{causal feature learning} (CFL) to derive macrovariables from microvariables in datasets with some minimal causal structure. For example, \cite{Chalupka:2016:UDE:3020948.3020957} analyze wind speeds and sea surface temperatures in a particular region of the Western Pacific Ocean. Using CFL, they partition both sets of variables into coarse-grained clusters that reveal the causal association between these fine-grained observations and large-scale weather patterns including El Ni\~no and La Ni\~na.\par

Chalupka et al.'s approach is related to work by \cite{hoel2013quantifying} and \cite{hoel2017map}. Although Hoel et al.'s formal approach (unlike Chalupka et al.'s) explicitly incorporates information theory, the approaches are similar in that they define an optimal coarse-graining for variables in a causal model using only the probabilistic relationships between variables. Additionally, work by \cite{beckers2019approximate} builds on work by \cite{Rubensteinetal17} and \cite{beckers2019abstracting} to define a scheme for coarse-graining causal variables such that the resulting causal graph is an approximation, to some degree of precision, of the underlying, fine-grained graph. While what we present here is in broad agreement with the spirit of this approach, our proposal is different in that it explicitly represents the extent to which the optimal level of approximation can be determined by the pragmatic interests of an agent.\par

The merits of CFL notwithstanding, we argue that the method is ill-equipped to handle the variable interests of real-world agents who may undertake causal analyses of a target system with different goals in mind. We demonstrate that by failing to incorporate pragmatic information, CFL is prone to errors in both directions -- failing to cluster values of a causal variable that should be grouped together, and clustering values of an effect variable that should be kept separate. We present an algorithm for \textit{pragmatic causal feature learning} (PCFL) that avoids these pitfalls without sacrificing any of the measure-theoretic advantages of the original CFL method.\par 

The remainder of this paper is structured as follows. In Sect.~\ref{sec:cfl}, we outline the original CFL algorithm. We present two examples in Sect.~\ref{sec:probz} that demonstrate how CFL can generate errors even in relatively simple cases. We introduce PCFL in Sect.~\ref{sec:pcfl}, outline an algorithm for implementing the procedure in Sect.~\ref{sec:pcfl_alg}, and present experimental results in Sect.~\ref{sec:exps}. Sect.~\ref{sec:conclusion} concludes.\par 

\section{Causal Feature Learning}\label{sec:cfl}
Let $(C,E)$ be a pair of discretely-valued random variables in a set $\mathcal{V}$. Let $(\mathcal{V},\mathcal{E},p(\cdot))$ be a Bayesian network in which $\mathcal{E}$ is an acyclic set of ordered pairs (or directed edges) of variables in $\mathcal{V}$, and $p(\cdot)$ is a joint probability distribution over the cross-product of the ranges of the variables in $\mathcal{V}$. In the graph $\mathcal{G}=(\mathcal{V},\mathcal{E})$, $E$ is a descendant of $C$. In keeping with the theory of Bayesian networks developed by \cite{pearl1988probabilistic}, the graph $(\mathcal{V},\mathcal{E})$ is Markov to the probability distribution $p(\cdot)$. Let $p(e_{i}|\hat{c}_{j})$ be an \textit{interventional conditional probability}, in which the notation $\hat{c}_{j}$ indicates that the value of $C$ has been set via an exogenous intervention on the system represented by the Bayesian network $(\mathcal{V},\mathcal{E},p(\cdot))$. Let $R_{X}$ be the range of a random variable $X$. Finally, a set $A$ is a \textit{quotient set} of a set $B$ iff each element of $A$ is an equivalence class of elements of $B$, according to some equivalence relation $\sim$.\par 

Chalupka et al.'s proposal for coarsening variables in causal models is straightforward. First, they define equivalence relations over the ranges of cause and effect variables. These equivalence relations form the basis of a coarsening process that generates macrovariables whose values are just the equivalence classes of the microvariables. Chalupka et al.\ define the equivalence relation $\sim_{c}$ over the range $R_{C}$ as follows:
\blockquote{\textbf{Causal Equivalence}:\ $c_{j}\sim_{c} c_{k}$ with respect to $E$ iff $p(e_{i}|\hat{c}_{j})=p(e_{i}|\hat{c}_{k})$ $\forall e_{i}\in R_{E}$.}
Let $C^{\dagger[E]}$ be the \emph{causal coarsening} of $C$ with respect to $E$ iff its range $R_{C^{\dagger[E]}}$ is the quotient set of $R_{C}$ induced by the equivalence relation $\sim_{c}$ with respect to $E$. Chalupka et al.'s equivalence relation $\sim_{e}$ over the range $R_{E}$ is defined as follows:
\blockquote{\textbf{Effect Equivalence}:\ $e_{i}\sim_{e}e_{s}$ with respect to $C$ iff $p(e_{i}|\hat{c}_{k})=p(e_{s}|\hat{c}_{k})$ $\forall c_{k}\in R_{C}$.}
Let $E^{*[C]}$ be the \emph{effect coarsening} of $E$ with respect to $C$ iff its range $R_{E^{*[C]}}$ is the quotient set of $R_{E}$ induced by the equivalence relation $\sim_{e}$ with respect to $C$.\par

Chalupka et al.'s principal formal achievement is to prove that, except in a Lebesgue measure zero subset of cases, the causal coarsening and the effect coarsening of a given variable can be learned from observational rather than experimental data. To state their result in a perspicuous way, we first define additional equivalence relations and coarsenings over the variables in the cause-effect pair $(C,E)$. Let us begin with the \textit{observational causal equivalence} relation $\sim_{oc}$, which is defined as follows:
\blockquote{\textbf{Observational Causal Equivalence}: $c_{j}\sim_{oc} c_{k}$ with respect to $E$ iff $p(e_{i}|c_{j})=p(e_{i}|c_{k})$ $\forall e_{i}\in R_{E}$.}
Note that the sole difference between observational causal equivalence and causal equivalence \textit{simpliciter} is that the former is defined using observational conditional probabilities (i.e., conditional probabilities such that the conditioning event is \textit{not} set via an intervention), whereas the latter is defined using causal conditional probabilities (i.e., conditional probabilities such that the conditioning event \textit{is} set via an intervention). Let $C^{oc[E]}$ be the \emph{observational causal coarsening} of $C$ with respect to $E$ iff the range $R_{C^{oc[E]}}$ is the quotient set of the range $R_{C}$ induced by the equivalence relation $\sim_{oc}$ with respect to $E$. Similar definitions of an observational equivalence relation and accompanying coarsening can be given for the effect variable:
\blockquote{\textbf{Observational Effect Equivalence}: $e_{i}\sim_{oe} e_{s}$ with respect to $C$ iff $p(e_{i}|c_{k})=p(e_{s}|c_{k})$ $\forall c_{k}\in R_{C}$.}
Let $E^{oe[C]}$ be the \emph{observational effect coarsening} of $E$ with respect to $C$ iff the range of $E^{oe[C]}$ is a quotient set of the range of $E$ according to the equivalence relation $\sim_{oe}$ with respect to $C$.\par

The distributions $p(E|c_j)$ and $p(E|\hat{c}_j)$ are not necessarily equivalent. For instance, the two may diverge due to some confounding variable $Z$ that is a cause of both $C$ and $E$. Alternatively, $Z$ may be an effect of $C$ and a cause of $E$. In both cases, there may be differences between observational and interventional distributions, such that the observational coarsenings will differ from causal coarsenings of both the cause and effect variables. Chalupka et al.\ argue that when we coarsen fine-grained cause and effect variables, we can almost always ignore the distorting influence of potential confounders.\par

\begin{proposition}[Chalupka et al.\ \citeyear{chalupka2017causal}, p.\ 149]\label{cctprop}
Let $C$, $E$, and $Z$ be variables in a graph $\mathcal{G}$ such that $C$ is an ancestor of $E$ and $Z$ is a possible confounder of the causal relationship between $C$ and $E$. Consider the set of possible joint probability distributions over $C$, $E$, and $Z$. Let $C^{\dagger[E]}$ and $C^{oc[E]}$ be the causal coarsening and observational causal coarsening of $C$, respectively, in any such probability distribution. Let $E^{*[C]}$ and $E^{oe[C]}$ be the effect coarsening and observational effect coarsening of $E$, respectively, in any such probability distribution. The set of joint probability distributions over $C$, $E$, and $Z$ such that the range of $C^{\dagger[E]}$ is not a quotient set of the range of $C^{oc[E]}$ and the range of $E^{*[C]}$ is not a quotient set of the range of $E^{oe[C]}$ is Lebesgue measure zero within the set of all possible joint distributions over $C$, $E$, and $Z$.  
\end{proposition}
\noindent
This result has a putatively important practical consequence. Suppose that we want to find the causal coarsening of some fine-grained variable $C$ that is a cause of some other fine-grained variable $E$. If we are guided solely by the coarsening strategies outlined above, then doing so would require a separate intervention for each value of $C$, since each causal conditional probability would be needed to determine which values of $C$ and $E$ are equivalent. However, equipped with Proposition \ref{cctprop}, we can instead use a potentially more efficient procedure. Using just observational data, we can coarsen the microvariable $C$ into the observational macrovariable $C^{oc[E]}$. We can then intervene to set $C^{oc[E]}$ to each of its values, and thereby determine which (if any) are equivalent according to the relation $\sim_{c}$ with respect to $E$, thereby generating the causal coarsening $C^{\dagger[E]}$. This procedure is potentially more efficient since, by construction, $|R_{C^{oc[E]}}| \leq |R_C|$. The procedure is justified just in case the joint probability distribution over $C$, $E$, and $Z$ is such that $C^{\dagger[E]}$ is a coarsening of $C^{oc[E]}$, a condition that provably only fails for a measure zero set of probability distributions. An analogous argument applies to the effect variable $E$.\par

\section{Problems for Causal Feature Learning}\label{sec:probz}

\begin{table}
\RawFloats
\footnotesize
\centering
\parbox{.45\linewidth}{
\centering
 \begin{tabular}{|| c | c | c ||}
 \hline
\ & $\texttt{[0,49]}$ & $\texttt{[50,69]}$\\ [0.5ex] 
 \hline
\texttt{Marlboro} & .026 & .25\\ 
 \hline
\texttt{Other} & .024 & .25\\
\hline
\texttt{Nothing} & .001 & .05\\
\hline
\ & $\texttt{[70,90]}$ & $\texttt{[90,\text{Inf}]}$\\ [0.5ex] 
 \hline
\texttt{Marlboro} & .698 & .026 \\ 
 \hline
\texttt{Other} & .702 & .024\\
\hline
\texttt{Nothing} & .948 & .001\\
\hline
\end{tabular}
\caption{Causal CPT}\label{90plus}}
\hfill
\parbox{.45\linewidth}{
\centering
 \begin{tabular}{|| c | c | c ||}
 \hline
\ & $\texttt{[0,49]}$ & $\texttt{[50,69]}$\\[0.5ex] 
 \hline
\texttt{Marlboro} & -950 & 1100\\ 
 \hline
\texttt{Other} & -990 & 1050\\
\hline
\texttt{Nothing} & -1000 & 1000\\
\hline
\ & $\texttt{[70,90]}$ & $\texttt{[90,Inf]}$\\[0.5ex]
 \hline
\texttt{Marlboro} & 2100 & 2150 \\ 
 \hline
\texttt{Other} & 2050 & 2145\\
\hline
\texttt{Nothing} & 2000 & 2050\\
\hline
\end{tabular}
\caption{Utilities for Smoking Decision}\label{utility}}
\end{table}

Suppose that a person's mortality age in years is represented by an effect variable $D$ such that $R_D = \{\texttt{[0,49], [50,69], [70,89], [90,Inf]}\}$. The causal variable $S$ describes the smoking habits of individuals, with  $R_S = \{\texttt{Marlboro, Other, Nothing}\}$. All causal conditional probabilities are given in Table~\ref{90plus}. In this example, the probability distribution over the effect variable given an intervention making someone a Marlboro smoker is very similar to the probability distribution over the effect variable given an intervention making someone a smoker of other brands. For many practical purposes, these two values of the causal variable are equivalent, and only trivially non-equivalent, such that the value space of the causal variable should be coarsened into the two-element set $\{\texttt{Smoker, Non-Smoker}\}$. However, CFL recommends against such a coarsening and therefore fails to satisfy an intuitive desideratum for any coarsening procedure -- namely, that it groups together causal values that are only trivially non-equivalent in a given context.\par 

CFL can also lead to counterintuitive results for effect variable coarsening. Consider again the conditional probability distribution in Table~\ref{90plus}. On Chalupka et al.'s account, the values \texttt{[0,49]} and \texttt{[90,Inf]} should be coarsened into a single discontinuous value \texttt{[0,49]} $\vee$ \texttt{[90,Inf]}. However, this coarse-grained value does not seem to pick out any meaningful scientific category. After all, dying before age fifty and dying after age ninety are very different outcomes from both an ontological and a pragmatic standpoint. Yet these outcomes are equivalent according to Chalupka et al.'s definition. Thus, CFL fails to satisfy another desideratum of a causal coarsening procedure (viz., that it not group together effect values that are only trivially equivalent in a given context).\par

\section{Pragmatic Causal Feature Learning}\label{sec:pcfl}
In this section, we present a novel approach to CFL wherein values of the cause and effect variables are coarsened together iff doing so does not decrease the maximum expected utility of intervening on the causal variable for an agent with a pragmatic interest in the target system. To formalize this idea, let $u(\cdot):R_{C}\times R_{E}\rightarrow\mathbbm{R}$ be a function that represents the utility some agent receives when a given pair of cause and effect values obtains.\footnote{This utility function can be obtained from data regarding an agent's preferences over cause-effect pairs; see \cite{morgenstern1953theory}.} For each value $c_{j}$ of the causal variable, we define the following set of utilities $U_{E|c_{j}}=\{u(c_{j},e_{1}),\dots,u(c_{j},e_{n})\}$ and set of  interventional conditional probabilities $P_{E|\hat{c}_{j}}=\{p(e_{1}|\hat{c}_{j}),\dots,p(e_{n}|\hat{c}_{j})\}$. The expected utility of setting causal variable $C$ to value $c_j$ via an intervention, for an agent with a utility function defined over the product space $R_{C}\times R_{E}$, is given by the inner product of these two sets. Let $\eta_{u}(C,E)=\max_{c_{j}\in R_{C}} \langle U_{E|c_{j}}, P_{E|\hat{c}_{j}}\rangle$ denote this agent's maximum expected utility over possible causal interventions. To illustrate, consider an agent who is deliberating whether to smoke cigarettes, and which brand to smoke. The agent's payoffs are given in Table~\ref{utility}. Using the interventional conditional probability table shown in Table~\ref{90plus}, we calculate that if $S$ is a variable such that the possible interventions on $S$ are the smoker statuses listed in the rows of Table~\ref{utility}, and the four death age categories in the columns of Table~\ref{utility} comprise the range of a variable $D$, then $\eta_{u}(S,D)\approx1947.05$. In what follows, we define equivalence relations over fine-grained cause and effect variables such that the corresponding coarsenings maximize $\eta_{u}(\cdot)$.\par

We define a pragmatic equivalence relation $\sim_{pc}$ between two values of the causal variable:\ 
\blockquote{\textbf{Pragmatic Causal Equivalence}: $c_{j} \sim_{pc} c_{k}$ with respect to $E$ iff either of the following holds:\ i) $\langle U_{E|c_{j}}, P_{E|\hat{c}_{j}} \rangle$ $= \langle U_{E|c_{k}}, P_{E|\hat{c}_{k}} \rangle$ $= \eta_{u}(C,E)$,
ii) $\langle U_{E|c_{j}}, P_{E|\hat{c}_{j}} \rangle$ $\neq \eta_{u}(C,E)$ and $\langle U_{E|c_{k}}, P_{E|\hat{c}_{k}} \rangle$ $\neq \eta_{u}(C,E)$.}
$C^{pc[E]}$ is the \emph{pragmatic causal coarsening} of $C$ with respect to $E$ iff its range $R_{C^{pc[E]}}$ is the quotient set of $R_{C}$ induced by the equivalence relation $\sim_{pc}$ with respect to $E$. By construction, $C^{pc[E]}$ is equivalent to a Boolean variable that takes a value of 1 for all and only those fine-grained $c_j$ that maximize an agent's expected utility according to some function $u(\cdot)$. All other values of $C$ correspond to sub-optimal interventions from the point of view of an agent who aims to maximize expected utility. Using Tables \ref{90plus} and \ref{utility}, one can verify that \texttt{Marlboro} and \texttt{Other} are pragmatic causal equivalents. Thus, PCFL successfully coarsens together these two values of $C$, whereas CFL fails to do so. We define a pragmatic equivalence relation $\sim_{pe}$ between two values of the effect variable:\ 
\blockquote{\textbf{Pragmatic Effect Equivalence}:\ $e_{i}\sim_{pe}e_{s}$ with respect to $C$ iff $u(c_{k},e_{i})=u(c_{k},e_{s})$ $\forall c_{k}\in R_{C}$.}
$E^{pe[C]}$ is the \emph{pragmatic effect coarsening} of $E$ with respect to $C$ iff its range $R_{E^{pe[C]}}$ is the quotient set of $R_{E}$ induced by the equivalence relation $\sim_{pe}$ with respect to $C$. Using Table~\ref{utility}, one can verify that \texttt{[0,49]} and \texttt{[90,Inf]} are \textit{not} pragmatic effect equivalents. Thus, PCFL successfully refrains from coarsening together these two values of $E$, whereas CFL fails to do so.\par

We show that PCFL has the same measure-theoretic advantages as CFL. First, we define the following \textit{observational} pragmatic equivalence relation over $R_{C}$:
\blockquote{\textbf{Observational Pragmatic Causal Equivalence}:\ $c_{j}\sim_{opc}c_{k}$ with respect to $E$ iff $\langle U_{E|c_{j}}, P_{E|c_{j}} \rangle = \langle U_{E|c_{k}}, P_{E|c_{k}} \rangle$.}
There are two salient differences between observational pragmatic causal equivalence and pragmatic causal equivalence \textit{simpliciter}. First, the observational pragmatic causal equivalence is calculated using conditional probability sets of the form $P_{E|c_{j}}=\{p(e_{1}|c_{j}),\dots,p(e_{n}|c_{j})\}$, which contains observational conditional probabilities rather than interventional conditional probabilities. Second, observational pragmatic causal equivalence is not assessed relative to any maximum expected utility; we only check whether expected utility is the same, given an observation of each value of $C$.\par

Let $C^{opc[E]}$ be the \emph{observational pragmatic causal coarsening} of $C$ with respect to $E$ iff its range $R_{C^{opc[E]}}$ is the quotient set of $R_{C}$ induced by the equivalence relation $\sim_{opc}$ with respect to $E$. We can now state the following proposition:
\begin{proposition}\label{pcctprop}
Let $C$, $E$, and $Z$ be variables in a graph $\mathcal{G}$ such that $C$ is an ancestor of $E$, and $Z$ is a possible confounder of the causal relationship between $C$ and $E$. Consider the set of possible joint probability distributions over $C$, $E$, and $Z$. Let $C^{pc[E]}$ and $C^{opc[E]}$ be the pragmatic causal coarsening and observational pragmatic causal coarsening of $C$, respectively, in any such probability distribution. The set of joint probability distributions over $C$, $E$, and $Z$ such that the range of $C^{pc[E]}$ is not a quotient set of the range of $C^{opc[E]}$ is Lebesgue measure zero within the set of all possible joint distributions over $C$, $E$, and $Z$.\footnote{Note that both this result and Prop.~\ref{cctprop} are only significant if one accepts that the distribution over the set of possible distributions over variables is given by the Lebesgue measure. Thus, we claim only that our approach is on equal footing with Chalupka et al.'s with respect to the putatively good-making features of an algorithm established by these geometric results.}
\end{proposition}
\noindent
There is no formal analog of the causal coarsening theorem for pragmatic effect coarsening. However, it is still the case that we can efficiently learn the pragmatic coarsening of an effect variable from just the observational conditional probability distribution over the causal variable, with failure only occurring in a Lebesgue measure zero subset of probability distributions. This is because the pragmatic effect equivalence relation holds or does not hold between two values of an effect variable independently of the probability distribution over the cause-effect pair $(C,E)$. Thus, we can use the following procedure to learn the pragmatic coarsening of a pair $(C,E)$ from observational data. First, obtain the observational pragmatic coarsening $C^{opc[E]}$. Next, intervene on each value of $c^{opc[E]}_{j}\in R_{C^{opc[E]}}$ and calculate $\langle U_{E|\hat{c}^{opc[E]}_{j}}, P_{E|\hat{c}^{opc[E]}_{j}} \rangle$ to obtain the pragmatic causal coarsening $C^{pc[E]}$. This procedure will only fail to obtain the true range of $C^{pc[E]}$ on a Lebesgue measure zero subset of probability distributions over $(C,E)$. Finally, use the utility function over $R_{C}\times R_{E}$ to determine which values of the effect variable are pragmatic effect equivalents in order to obtain the pragmatically coarsened cause-effect pair $(C^{pc[E]},E^{pe[C]})$.\par

\section{An Algorithm for Pragmatic Causal Feature Learning}\label{sec:pcfl_alg}

\begin{figure*}[ttt!]
\begin{minipage}{.48\linewidth}
\begin{algorithm}[H]
\footnotesize
\caption{CFL}\label{chalupkaalgo}
\begin{algorithmic}
\item \textbf{input} : $\mathcal{D}=\{(c_{1},e_{1}),\dots,(c_{N},e_{N})\}$ \\ \ \ \ \ \ \ \ \ \ \ \texttt{Cluster} - \text{a clustering algorithm}

\item \textbf{output} : $W(c_{i})$, $T(e_{i})$ 
\end{algorithmic}

\begin{algorithmic}[1]
\item Regress $f\leftarrow\argmin_{f}\sum_{1}^{N}(f(c_{i})-e_{i})^{2}$;
\item Let $W(c_{i})\leftarrow\texttt{Cluster}(f(c_{1}),\dots,f(c_{N}))[c_{i}]$;
\item Let $\texttt{Range}(W)=\{c^{oc[E]}_{1},\dots,c^{oc[E]}_{\upsilon}\}$;
\item Let $\mathfrak{E}_{\beta}\leftarrow\{e_{i}|W(c_{i})=c^{oc[E]}_{\beta} \ \text{and} \ (c_{i},e_{i})\in\mathcal{D}\}$;
\item Let $g(e_{i})\leftarrow[\texttt{kNN}(e_{i},\mathfrak{E}_{\beta}),\dots,\texttt{kNN}(e_{i},\mathfrak{E}_{\upsilon})]$;
\item Let $T(e_{i})\leftarrow\texttt{Cluster}(g(e_{1}),\dots,g(e_{N}))[e_{i}]$;
\end{algorithmic}
\end{algorithm}
\end{minipage}
\hfill
\begin{minipage}{.48\linewidth}
\begin{algorithm}[H]
\footnotesize
\caption{PCFL}\label{pragalgo}
\begin{algorithmic}
\item \textbf{input} : $\mathcal{D}=\{(c_{1},e_{1}),\dots,(c_{N},e_{N})\}$\\ 
\ \ \ \ \ \ \ \ \ \ \  $\mathcal{U}=\{u(c_{1},e_{1}),\dots,u(c_{N},e_{N})\}$\\
\ \ \ \ \ \ \ \ \ \ \  \texttt{Cluster} - \text{a clustering algorithm}

\item \textbf{output} : $W_{p}(c_{i})$, $T_{p}(e_{i})$ 
\end{algorithmic}

\begin{algorithmic}[1]
\item Regress $f\leftarrow\argmin_{f}\sum_{1}^{N}(f(c_{i})-u(c_{i},e_{i}))^{2}$;
\item Let $W_{p}(c_{i})\leftarrow\texttt{Cluster}(f(c_{1}),\dots,f(c_{N}))[c_{i}]$;
\item Let $g(e_{i})\leftarrow[u(c_{1},e_{1}),\dots,u(c_{N},e_{N})]$;
\item Let $T_{p}(e_{i})\leftarrow\texttt{Cluster}(g(e_{1}),\dots,g(e_{N}))[e_{i}]$;
\end{algorithmic}
\end{algorithm}
\end{minipage}
\end{figure*}

Alg. \ref{chalupkaalgo} presents \cite{Chalupka:2016:UDE:3020948.3020957}'s original CFL algorithm, rephrased to conform to the formalisms used in this paper. Note that this algorithm only finds the \textit{observational} coarsenings of the fine-grained cause and effect variable. Thus, it serves mainly to compress the space of interventions needed to learn the full causal coarsenings $C^{\dagger[E]}$ and $E^{*[C]}$, under the assumption that the interventional conditional probability distribution over $E$, given each intervention on $C$, is not known and cannot be learned via inference from the true causal structure (i.e., the true causal structure is also not known). The algorithm assumes that the fine-grained variables $C$ and $E$ can take a finite number of continuous or categorical values, and takes as input $N$ observations of values for $C$ and $E$. Note that line 1 regresses $E$ on $C$ with $L_2$ loss, thereby learning the conditional expectation $\mathbbm{E}[E|C]$. In general $\mathbbm{E}[E|c_{j}]=\mathbbm{E}[E|c_{k}]$ is a necessary but insufficient condition for $c_{j}\sim_{oc} c_{k}$ with respect to $E$. The mean value of a random variable can be identical under two distributions even if those distributions are not identical. However, Chalupka et al.\ use $\mathbbm{E}[E|c_{j}]=\mathbbm{E}[E|c_{k}]$ as a ``heuristic indicator'' for $c_{j}\sim_{oc} c_{k}$ with respect to $E$, while accepting that this heuristic may sometimes fail (\citeyear{Chalupka:2016:UDE:3020948.3020957}, p.\ 6). Note also that the algorithm \texttt{kNN} in line 5 returns the distance between $e_{i}$ and its $k$-th nearest neighbor in the set constructed in Line 4, for some arbitrarily selected $k$.\footnote{Chalupka et al. use Euclidean distance, but other measures could of course be substituted here.} If two values $e_{i}$ and $e_{s}$ of $E$ are the same distance from their $k$-th nearest neighbor in each of the sets constructed in Line 4, such that they will be clustered together in line 6, then $p(e_{i}|c^{oc[E]}_{\beta})=p(e_{s}|c^{oc[E]}_{\beta})$ for all $c^{oc[E]}_{\beta}$, as shown by \cite{fukunaga1973optimization}.\par 

Alg. \ref{pragalgo} provides pseudo-code for PCFL. Crucially, this algorithm takes as an additional input the utilities that the agent assigns to each observed cause-effect pair. Note that Line 1 regresses $\mathcal{U}$ on $C$ with $L_2$ loss, thereby finding the conditional expected utility $\mathbbm{E}[u(c_{i},E)|c_{i}]$ for each $c_{i}$. Since $\mathbbm{E}[u(c_{i},E)|c_{i}] = \langle U_{E|c_{i}}, P_{E|c_{i}} \rangle$, the regression in line 1 allows us to define, in Line 2, a function $W_{p}(\cdot)$ whose range consists of all and only those values of the observational pragmatic causal coarsening $C^{opc[E]}$ that appear in the dataset $\mathcal{D}$. Unlike in CFL, the expectations $\mathbbm{E}[u(c_{i},E)|c_{i}]$ are not heuristic indicators of an equivalence relationship, the use of which may lead to inaccurate output. Rather, we can use these expectations to directly learn which values of $C$ are observational pragmatic causal equivalents with respect to $E$.\par

\section{Experimental Results}\label{sec:exps}
\subsection{Simulated Data}\label{sec:scm}
We implement Alg.\ \ref{chalupkaalgo} and Alg.\ \ref{pragalgo} on a simulated dataset to compare the performance of both methods.\footnote{Code for all simulations and algorithms is available at \texttt{https://github.com/davidbkinney/pcfl}.} Consider a Bayes net $M$ (see Fig.~\ref{fig:scm}) with two unobserved binary variables $Z_{1} \sim \text{Bern}(0.5)$ and $Z_{2} \sim \text{Bern}(0.5)$. Let $R_C = R_E = \{-2, -1, 1, 2\}$. Values of the causal variable $C$ are fixed by the exogenous variables as follows:\ $C=-2$ if $Z_{1}=0$ and $Z_{2}=0$, $C=-1$ if $Z_{1}=0$ and $Z_{2}=1$, $C=1$  and $Z_{1}=1$ and $Z_{2}=0$, and $C=2$ if $Z_{1}=1$ and $Z_{2}=1$. Conditional probabilities for the effect variable $E$ are given by the equation $P(E|C,Z_1) = \sigma(\bm{\alpha} + C \bm{\beta} + Z_{1} \bm{\gamma})$, where $\sigma(\cdot)$ denotes the softmax transformation and linear parameters are computed to induce the conditional probabilities listed in Table~\ref{condsim}. Note that the exogenous variable $Z_1$ has directed edges into both $C$ and $E$, thereby confounding our causal relationship of interest, $C \to E$. It is clear from Table~\ref{condsim} that $C=-1$ and $C=1$ are observational causal equivalents, while $E=-2$ and $E=2$ are observational effect equivalents, according to the definitions in Sect.~\ref{sec:cfl}. Thus, we expect CFL’s coarsened conditional probability table to appear as it does in Table~\ref{expectedsim}. This is empirically verified by a simulation experiment in which we run CFL on 10,000 samples drawn from $P(M)$. Resulting probabilities are reported in Table~\ref{chalpukaresultssim}.

\begin{figure}

\begin{floatrow}
\ffigbox{%
 \begin{tikzpicture}[scale=1, node distance = .8cm]
    \node[] (1) {};
    \node[right =of 1] (2) {$Z_{1}$};
    \node[right =of 2] (3) {};
    \node[below =of 1] (4) {$Z_{2}$};
    \node[right =of 4] (5) {$C$};
    \node[right =of 5] (6) {$E$};
    \draw[->] (5) to (6);
    \draw[dashed,->] (4) to (5);
    \draw[dashed,->] (2) to (5);
    \draw[dashed,->] (2) to (6);
    \end{tikzpicture}}
    {\caption{Causal diagram for the graph $M$.}
    \label{fig:scm}
}
\capbtabbox[\linewidth]{%
\footnotesize
 \begin{tabular}{|| c | c | c | c | c ||}
 \hline
       & $E=-2$ & $E=-1$ & $E=1$  & $E=2$ \\
\hline
$C=-2$ & $.248$ & $.189$ & $.315$ & $.248$ \\[.3ex]
\hline
$C=-1$ & $.252$ & $.248$ & $.248$ & $.252$ \\[.3ex]
\hline
$C=1$  & $.252$ & $.248$ & $.248$ & $.252$ \\[.3ex]
\hline
$C=2$  & $.248$ & $.315$ & $.189$ & $.248$ \\[.3ex]
\hline
\end{tabular}}{%
  \caption{Expected conditional probabilities, with expectation taken over the graph $M$.}
    \label{condsim} 
    }
\end{floatrow}
\end{figure}

\begin{table}

\RawFloats
\footnotesize
	\centering
\parbox{.48\linewidth}{
	\centering
 \begin{tabular}{|| c | c | c | c ||}
 \hline
       & $E=-2\vee2$ & $E=-1$ & $E=1$ \\
\hline
$C=-2$      & $.496$ & $.189$ & $.315$ \\[.3ex]
\hline
$C=-1\vee1$ & $.504$ & $.248$ & $.248$ \\[.3ex]
\hline
$C=2$       & $.496$ & $.315$ & $.189$ \\[.3ex]
\hline
\end{tabular}
\caption{Expected output of CFL algorithm on data sampled from $M$.}\label{expectedsim}}
\hfill
\parbox{.48\linewidth}{
	\centering
 \begin{tabular}{|| c | c | c | c ||}
 \hline
 & $E=-2\vee2$ & $E=-1$ & $E=1$ \\
\hline
$C=-2$ & $.491$ & $.186$ & $.323$ \\[.3ex]
\hline
$C=-1\vee1$ & $.510$ & $.255$ & $.236$ \\[.3ex]
\hline
$C=2$ & $.514$ & $.295$ & $.191$ \\[.3ex]
\hline
\end{tabular}
\footnotesize
\caption{Observed output of CFL algorithm on data sampled from $M$.}\label{chalpukaresultssim}}
\end{table}

\begin{table}

\footnotesize
	\centering
 \begin{tabular}{|| c | c | c | c | c ||}
 \hline
       & $E=-2$ & $E=-1$ & $E=1$ & $E=2$ \\
\hline
$C=-2$ & 1      & 2      & 2     & 4 \\
\hline
$C=-1$ & 8      & 5      & 5     & 0 \\
\hline
$C=1$  & 5      & 8      & 8     & 9 \\
\hline
$C=2$  & 4      & 2      & 2     & 1 \\[.3ex]
\hline
\end{tabular}
\caption{Utility matrix for PCFL simulation experiment.}\label{utilsim}
\end{table}

To test the performance of Alg.\ \ref{pragalgo}, we introduce a utility matrix over cause-effect pairs (see Table~\ref{utilsim}). Using these utilities, along with the conditional probabilities in Table~\ref{condsim}, one can calculate the expected utility of each $c_j \in R_C$, and observe that $C=-2$ and $C=2$ are observational pragmatic causal equivalents; both have an expected utility of 2.25. By contrast, the expected utilities of $C=-1$ and $C=1$ are 4.50 and 7.50, respectively, and are therefore only observationally pragmatically equivalent to themselves. In addition, we observe that $E=-1$ and $E=1$ are observational pragmatic effect equivalents, while $E=-2$ and $E=2$ are only observationally pragmatically equivalent to themselves. On this basis, and retaining the assumption that all values of $C$ are equiprobable, we expect that PCFL will yield the coarsened conditional probabilities shown in Table~\ref{expectedpcflsim}. This is verified by our simulation test, where we obtain the correct pragmatic observational coarsening. Resulting probabilities are reported in Table~\ref{observedpcflsim}.

\begin{table}

\RawFloats
\footnotesize
\parbox{.45\linewidth}{
 \begin{tabular}{|| c | c | c | c ||}
 \hline
            & $E=-2$ & $E=-1\vee1$ & $E=2$ \\
\hline
$C=-2\vee2$ & $.248$ & $.504$      & $.248$ \\[.3ex]
\hline
$C=-1$      & $.252$ & $.496$       & $.252$ \\[.3ex]
\hline
$C=1$       & $.252$ & $.496$       & $.252$ \\[.3ex]
\hline
\end{tabular}
\caption{Expected output of PCFL algorithm on data sampled from $M$.}\label{expectedpcflsim}}
\hfill
\parbox{.45\linewidth}{
 \begin{tabular}{|| c | c | c | c ||}
 \hline
 & $E=-2$ & $E=-1\vee1$ & $E=2$ \\
\hline
$C=-2\vee 2$ & $.251$ & $.498$ & $.252$ \\[.3ex]
\hline
$C=-1$ & $.257$ & $.494$ & $.249$ \\[.3ex]
\hline
$C=1$ & $.261$ & $.487$ & $.252$ \\[.3ex]
\hline
\end{tabular}
\caption{Observed output of PCFL algorithm on data sampled from $M$.}\label{observedpcflsim}}
\end{table}

We argue that, in the context of the utility function implied by Table~\ref{utilsim}, the coarsening in Table~\ref{observedpcflsim} is a more appropriate high-level representation of the target system than the coarsening in Table~\ref{chalpukaresultssim}. This is because the former coarsens together those values of the cause and effect variables that represent equally desirable states of affairs, from the point of view of an agent whose preferences are represented by the utilities given in Table~\ref{utilsim}. The coarsening shown in Table~\ref{chalpukaresultssim} does not have a similar significance, and in some applications can lead to errors, as shown in Sect.~\ref{sec:probz}. Note that the distinction is not reducible to continuity assumptions, since an agent may have valid reasons to coarsen together nonadjacent regions of the feature space. For instance, it is common in statistics to distinguish between significant and insignificant results of two-tailed hypothesis tests, a procedure which naturally groups together the largest and smallest values of a test statistic. Similarly, PCFL clusters $C=-2$ and $C=2$ together in this simulation, in accordance with the given utilities.

\subsection{El Ni\~no Data}
\begin{figure}

\begin{floatrow}
\ffigbox{%
 \includegraphics[scale=.3]{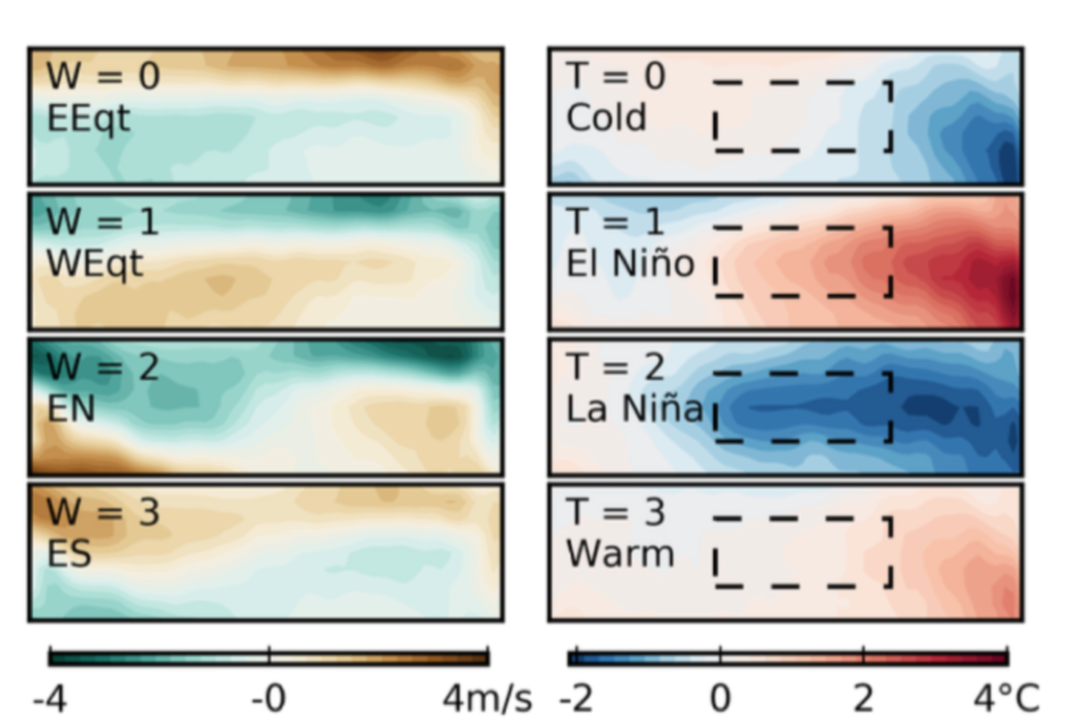}}
    {\caption{Alg.\ 1 output for the El Ni\~no dataset (\citeyear{Chalupka:2016:UDE:3020948.3020957}). Zonal wind fields are visualized left, sea surface temperatures right.}
    \label{chalupka_image}}
\ffigbox{%
 \includegraphics[scale=.25]{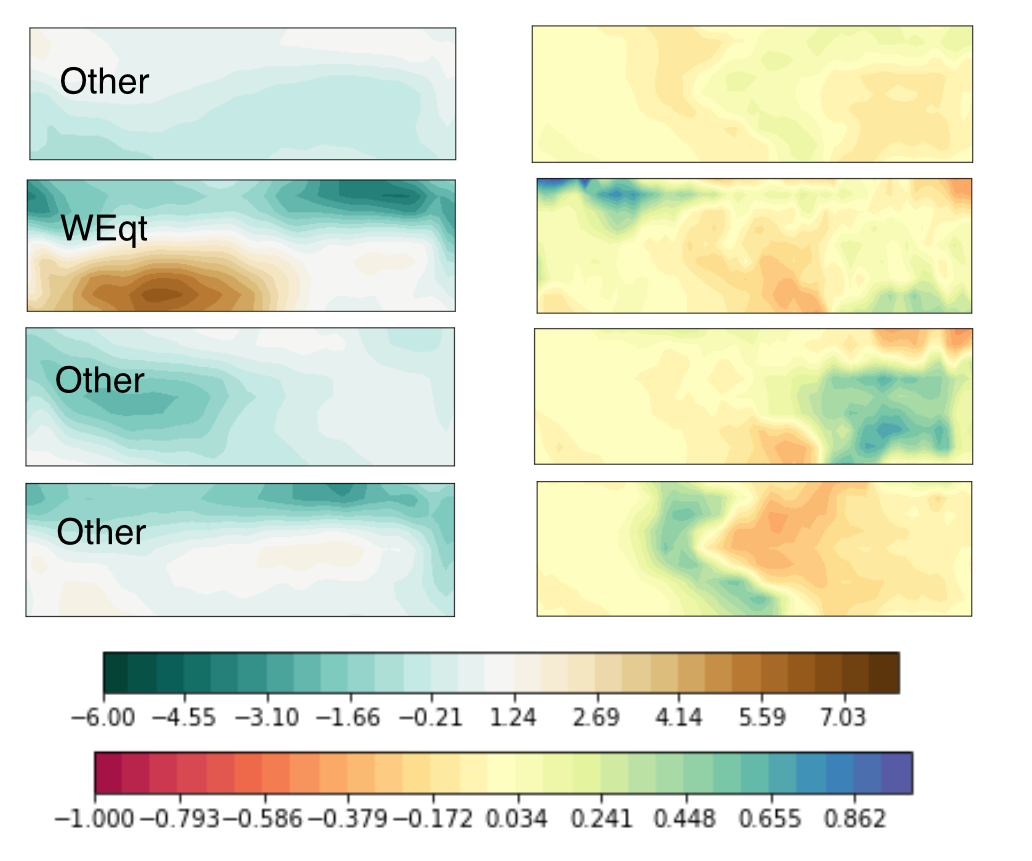}}
    {\caption{Alg.\ 2 output for the El Ni\~no dataset. Zonal wind fields are visualized left, sea surface temperatures right.}
    \label{my_image}}
\end{floatrow}
\end{figure}

We re-analyzed the dataset that \cite{Chalupka:2016:UDE:3020948.3020957} use in their study of El Ni\~no using both CFL and PCFL.  This dataset includes weekly average zonal wind speeds and sea surface temperatures from the same region of the Equatorial Pacific, each consisting of a $9 \times 55$ matrix of real numbers for each of 13,140 weeks. Fig. \ref{chalupka_image} shows the results of running CFL on this dataset; the algorithm picks out an Easterly Equatorial wind pattern, which causes slightly colder sea surface temperatures; a Westerly Equatorial wind pattern, which causes the warmer sea surface temperatures associated with the El Ni\~no effect; an Easterly North of the Equator wind pattern, which causes the colder temperatures associated with the La Ni\~na effect; and an Easterly South of the Equator wind pattern, which causes slightly warmer sea surface temperatures.\par

To implement PCFL, we let $X$ and $Y$ be the set of possible directional wind speeds and temperatures, respectively, in a given region of the Western Pacific. We then defined the following utility function over  $R_{X}\times R_{Y}$, where $y^{*}$ is the temperature rounded to the first decimal:\ 
\begin{equation}\label{eq:util}
    u(x,y) = -1 + \frac{1}{\sqrt{.02\pi}}e^{\frac{(y^{*}-26)^{2}}{.02}}
\end{equation}
Utilities for temperature observations are therefore determined by their distance from a mean of $26\degree$C using a radial basis function kernel with small bandwidth and a scale shift of $-1$. This encodes a strong preference on the part of the relevant agent for temperatures near $26\degree$C, rather than any observer-neutral value that this temperature might have. Note that in this case, the utility $u(x,y)$ is independent of the causal value $x$. As our earlier experiment on simulated data shows, our framework does not generally assume or require such independence. \par

Fig. \ref{my_image} shows the output of PCFL when it is set to discover four clusters of both $X$ and $Y$. The key difference between our results and those of Chalupka et al. is that whereas CFL picks out four trends in the data, the only distinct wind direction/speed pattern that PCFL identifies is an extreme version of the Westerly Equatorial wind pattern, shown in the second panel from the top. This causes mostly negative outcomes, as indicated in the pattern to the right of the Westerly Equatorial wind pattern; the darker orange regions represent highly negative outcomes, which are not offset by the small patches of dark-blue, positive outcomes in the upper left corner. Recalling that the Westerly Equatorial wind pattern is a cause of the El Ni\~no effect, our interpretation of this result is that an agent with the utility function defined by Eq. \ref{eq:util} is highly averse to the El Ni\~no weather pattern and relatively unconcerned about other meteorological phenomena. Thus, this agent picks out two main classes of wind patterns: the Westerly Equatorial pattern and other patterns.\par

\section{Discussion and Conclusion}\label{sec:conclusion}
The PCFL algorithm is a principled extension of Chalupka et al.'s CFL method. By incorporating pragmatic information about the inquiring agent's preferences over outcomes, we avoid certain counterintuitive and undesirable consequences of the original CFL algorithm without sacrificing its attractive measure theoretic properties. Our approach is also more computationally efficient if agents rely on utility functions less complex than the kNN algorithm. However, one could argue that this trivializes the problem of causal feature learning. If we are told what some agent cares about, then it seems that such an algorithm does not discover anything about the system under study, but instead regurgitates the interests of an arbitrary agent. In response, we argue that the dependence of our algorithm on an input utility function speaks to an important fact about the nature of scientific discovery. We maintain that the process of discovery, including the discovery of the salient possible macro-level states of a system from micro-level data, is a fundamentally goal-oriented process. A similar sentiment is echoed in work by \cite{Wellen2016-WELARL-2}, who argue that the interaction between agential goals and the environment is essential to understanding feature learning. Another framework that uses a similar formal apparatus to discuss coarsening, but coarsens variables according to different equivalence relations, is the framework of ``epsilon-machines for decisional states'' put forward by \cite{brodu2011reconstruction}. It would be a potentially fruitful extension of this paper to explore the connections between our framework and these approaches. In addition, \cite{beckers2019approximate} propose a more complex formalization of the relationship between coarse-grained and fine-grained models than the quotient-set relation used by both Chalupka et al. and ourselves, but do not consider the pragmatic context in which coarsening occurs. It would be fruitful to consider how the approach proposed here contrasts with their model.\par

\section*{Appendix:\ Proof of Proposition \ref{pcctprop}}
\begin{proof}
The proof closely follows the logic and methods of the first part of the proof of the causal coarsening theorem in \cite{chalupka2017causal}. We begin by introducing the following notation, for the purpose of concision: $i[i,l,j]=p(e_{i}|z_{l},c_{j})$, $\beta[j,l]=p(c_{j}|z_{l})$, $\gamma[l]=p(z_{l})$.
Note that $z_{l}$ is any value of a variable $Z$ that is a potential confounder of the relationship between $C$ and $E$, where $R_{Z}$ has cardinality $w$. Next, we define the following three vectors: $\vec{i}=[i[1,1,1],\dots,i[n,w,m]]$, $\vec{\beta}=[\beta[1,1],\dots,i[m,w]]$, $\vec{\gamma}=[\gamma[1],\dots,\gamma[w]]$. Each triple $(\vec{i},\vec{\beta},\vec{\gamma})$ is a point in a space $\mathbbm{R}^{d}$. The set of all such points forms a simplex $\mathcal{S}\subseteq\mathbbm{R}^{d}$. We want to show that the subset $\mathcal{S}^{\prime}\subseteq\mathcal{S}$ containing all joint distributions such that there are two $c_{j}$ and $c_{k}$ such that $c_{j}\sim_{opc}c_{k}$ with respect to $E$ but $c_{j}\not\sim_{pc}c_{k}$ with respect to $E$ is Lebesgue measure zero in $P[\vec{i},\vec{\beta},\vec{\gamma}]$, where $P[\vec{i},\vec{\beta},\vec{\gamma}]$ is the space of all possible points $(\vec{i},\vec{\beta},\vec{\gamma})$. To do this, we first fix $\vec{i}=\vec{i}^{*}$ and $\vec{\beta}=\vec{\beta}^{*}$, so that $\vec{\gamma}$ is the only free parameter. Let $P[\vec{\gamma};\vec{i}^{*},\vec{\beta}^{*}]$ be the set of all joint probability distributions consistent with this fixing of parameters. We proceed by showing that $\mathcal{S}^{\prime}\cap P[\vec{\gamma};\vec{i}^{*},\vec{\beta}^{*}]$ is Lebesgue measure zero in $P[\vec{\gamma};\vec{i}^{*},\vec{\beta}^{*}]$, and then integrating over all possible $\vec{i}$ and $\vec{\beta}$. To show that $\mathcal{S}^{\prime}\cap P[\vec{\gamma};\vec{i}^{*},\vec{\beta}^{*}]$ is Lebesgue measure zero in $P[\vec{\gamma};\vec{i}^{*},\vec{\beta}^{*}]$, pick any two values $c_{j}$ and $c_{k}$ such that $c_{j}\sim_{opc}c_{k}$ but $c_{j}\not\sim_{pc}c_{k}$. If no such pair exists, then we are done. If such a pair exists, then the fact that $c_{j}\sim_{opc}c_{k}$ means that the following constraint holds, for the fixed utility function $u(\cdot)$:
\begin{equation}\label{helper1}
    p(c_{j})^{-1}\sum_{i=1}^{n}\sum_{l=1}^{w}u(c_{j},e_{i})i^{*}[i,l,j]\beta^{*}[j,l]\gamma[l] = p(c_{k})^{-1}\sum_{i=1}^{n}\sum_{l=1}^{w}u(c_{k},e_{i})i^{*}[i,l,k]\beta^{*}[k,l]\gamma[l]
\end{equation}
We note that $p(c_{k})=\sum_{l=1}^{w}\beta^{*}[k,l]\gamma[l]$ and $p(c_{j})=\sum_{l=1}^{w}\beta^{*}[j,l]\gamma[l]$, which means that, after some algebra, (\ref{helper1}) implies:
\small
\begin{equation}\label{constraint}
\sum_{l^{\prime}=1}^{w}\sum_{l=1}^{w}\gamma[l^{\prime}]\gamma[l]\bigg(\sum_{i=1}^{n}u(c_{j},e_{i})i^{*}[i,l,j]\beta^{*}[k,l^{\prime}]\beta^{*}[j,l] \\ - \ \sum_{i=1}^{n}u(c_{k},e_{i})i^{*}[i,l,k]\beta^{*}[j,l^{\prime}]\beta^{*}[k,l]\bigg)=0
\end{equation}
\normalsize
Using an algebraic lemma from \cite{okamoto1973distinctness}, we know that $\mathcal{S}^{\prime}\cap P[\vec{\gamma};\vec{i}^{*},\vec{\beta}^{*}]$ is Lebesgue measure zero in $P[\vec{\gamma};\vec{i}^{*},\vec{\beta}^{*}]$ if this constraint is non-trivial, i.e.\ if there are $\vec{\gamma}$ such that (\ref{constraint}) does not hold. To show that this is the case, suppose first that all entries in $\vec{\gamma}$ equal $1/w$. If (\ref{constraint}) does not hold, then we are done. If (\ref{constraint}) does hold under this condition, then there are at least two pairs of values $(z_{q^{+}},z_{v^{+}})$ and $(z_{q^{-}},z_{v^{-}})$ such that the following hold:
\begin{equation}
    \sum_{i=1}^{n}u(c_{j},e_{i})i^{*}[i,q^{+},j]\beta^{*}[k,v^{+}]\beta^{*}[j,q^{+}] - \sum_{i=1}^{n}u(c_{k},e_{i})i^{*}[i,v^{+},k]\beta^{*}[j,q^{+}]\beta^{*}[k,v^{+}]>0
\end{equation}

\begin{equation}
    \sum_{i=1}^{n}u(c_{j},e_{i})i^{*}[i,q^{-},j]\beta^{*}[k,v^{-}]\beta^{*}[j,q^{-}] - \sum_{i=1}^{n}u(c_{k},e_{i})i^{*}[i,v^{-},k]\beta^{*}[j,q^{-}]\beta^{*}[k,v^{-}]<0
\end{equation}
This assumes that there is a $z_{l}$ such that (\ref{helper1}) does not hold. Indeed, if (\ref{helper1}) held, then it would not be the case that $c_{j}\sim_{opc}c_{k}$ with respect to $E$ but $c_{j}\not\sim_{pc}c_{k}$ with respect to $E$ (since the expected utility, given $c_{j}$ or $c_{k}$, would be the same regardless of the value of the confounder, such that there would be no difference between the interventional and observational probability distribution over $E$, given either value) and the proof would already be complete. The two inequalities above imply that either $z_{q^{+}}\neq z_{q^{-}}$, $z_{v^{+}}\neq z_{v^{-}}$, or both. Assume $z_{q^{+}}\neq z_{q^{-}}$, and pick any positive $\epsilon<\min\{1/w, 1 - 1/w\}$. For any $z_{l}$ such that $z_{l}\neq z_{q^{+}}$ and $z_{l}\neq z_{q^{-}}$, let $\gamma[l]=1/w$, while $\gamma[q^{+}]=1/w + \epsilon$ and $\gamma[q^{-}]=1/w - \epsilon$. This way, $\sum_{l^{\prime}=1}^{w}\sum_{l=1}^{w}\gamma[l^{\prime}]\gamma[l]$ is unchanged from the case where each entry in  $\vec{\gamma}$ is $1/w$, but it is nevertheless the case that (\ref{constraint}) does not hold. We can repeat these same steps under the assumption that $z_{v^{+}}\neq z_{v^{-}}$, and under the assumption that $z_{q^{+}}\neq z_{q^{-}}$ and $z_{v^{+}}\neq z_{v^{-}}$, generating the same result in each case. Since (\ref{constraint}) is non-trivial, we know that $\mathcal{S}^{\prime}\cap P[\vec{\gamma};\vec{i}^{*},\vec{\beta}^{*}]$ is Lebesgue measure zero in $P[\vec{\gamma};\vec{i}^{*},\vec{\beta}^{*}]$. We integrate over $\vec{i}$ and $\vec{\beta}$ to show that $\mathcal{S}^{\prime}$ is Lebesgue measure zero in $P[\vec{i},\vec{\beta},\vec{\gamma}]$. Let $\mathcal{S}^{\prime}=\cup_{\vec{i},\vec{\beta}}\tilde{P}[\vec{\gamma};\vec{i},\vec{\beta}]\subseteq P[\vec{i},\vec{\beta},\vec{\gamma}]$ be the Lebesgue measure zero set of all possible joint distributions $P[\vec{\gamma};\vec{i},\vec{\beta}]$ such that $c_{j}\sim_{opc}c_{k}$ with respect to $E$ but $c_{j}\not\sim_{pc}c_{k}$ with respect to $E$. We define a characteristic function $\theta$ such that $\theta(\vec{i},\vec{\beta},\vec{\gamma})=1$ if $\gamma\in\tilde{P}[\vec{\gamma};\vec{i},\vec{\beta}]$ and $\theta(\vec{i},\vec{\beta},\vec{\gamma})=0$ otherwise. By the basic properties of positive measures, we have $\mu(S^{\prime})=\int_{P[\vec{i},\vec{\beta},\vec{\gamma}]}\theta(\vec{i},\vec{\beta},\vec{\gamma})\ \text{d}\mu$. For concision, let $\mathcal{A}=\mathbbm{R}^{w\times n}$, let $\mathcal{B}=\mathbbm{R}^{n\times w}$, and let $\mathcal{G}=\mathbbm{R}^{w}$. We calculate $\mu(S^{\prime})$ as follows:
\begin{multline}
    \mu(S^{\prime})=\int_{\mathcal{A}\times \mathcal{B}\times\mathcal{G}}\theta(\vec{i},\vec{\beta},\vec{\gamma})\ \text{d}(\vec{i},\vec{\beta},\vec{\gamma}) = \int_{\mathcal{A}\times \mathcal{B}}\int_{\mathcal{G}}\theta(\vec{i},\vec{\beta},\vec{\gamma})\ \text{d}(\vec{\gamma}) \ \text{d}(\vec{i},\vec{\beta}) \\ = \ \int_{\mathcal{A}\times \mathcal{B}}\mu(\tilde{P}[\vec{\gamma};\vec{i},\vec{\beta}])\ \text{d}(\vec{i},\vec{\beta}) = \int_{\mathcal{A}\times \mathcal{B}}0\ \text{d}(\vec{i},\vec{\beta})=0
\end{multline}
Thus, the subset $\mathcal{S}^{\prime}\subseteq\mathcal{S}$ containing all joint distributions such that $c_{j}\sim_{opc}c_{k}$ but $c_{j}\not\sim_{pc}c_{k}$ is Lebesgue measure zero in $P[\vec{i},\vec{\beta},\vec{\gamma}]$.\end{proof}

\bibliography{pragbib}
\end{document}